\documentclass[conference,a4paper]{IEEEtran}
\IEEEoverridecommandlockouts

\usepackage{cite}
\usepackage{amsmath,amssymb,amsfonts}
\usepackage{algorithmic}
\usepackage{graphicx}
\usepackage{textcomp}
\usepackage{xcolor}
\usepackage{booktabs}
\usepackage{multirow}
\usepackage{url}
\usepackage[utf8]{inputenc}
\usepackage{array}

\def\BibTeX{{\rm B\kern-.05em{\sc i\kern-.025em b}\kern-.08em
    T\kern-.1667em\lower.7ex\hbox{E}\kern-.125emX}}

\begin{document}

\title{ArabicNumBench:  Evaluating Arabic Number Reading in Large Language Models}

\author{
\IEEEauthorblockN{Anas Alhumud}
\IEEEauthorblockA{\textit{Saudi Data and Artificial} \\
\textit{Intelligence Authority}\\
Saudi Arabia \\
ahumud@sdaia.gov.sa}
\and
\IEEEauthorblockN{Abdulaziz Alhammadi}
\IEEEauthorblockA{\textit{Imam Mohammad Ibn Saud} \\
\textit{Islamic University}\\
Saudi Arabia \\
446012270@sm.imamu.edu.sa}
\and
\IEEEauthorblockN{Muhammad Badruddin Khan}
\IEEEauthorblockA{\textit{Imam Mohammad Ibn Saud} \\
\textit{Islamic University}\\
Saudi Arabia \\
mbkhan@imamu.edu.sa}
}

\maketitle

\begin{abstract}
We present ArabicNumBench, a comprehensive benchmark for evaluating large language models on Arabic number reading tasks across Eastern Arabic-Indic numerals (0-9 in Arabic script) and Western Arabic numerals (0-9). We evaluate 71 models from 10 providers using four prompting strategies (zero-shot, zero-shot CoT, few-shot, few-shot CoT) on 210 number reading tasks spanning six contextual categories: pure numerals, addresses, dates, quantities, and prices. Our evaluation comprises 59,010 individual test cases and tracks extraction methods to measure structured output generation. Evaluation reveals substantial performance variation, with accuracy ranging from 14.29\% to 99.05\% across models and strategies. Few-shot Chain-of-Thought prompting achieves 2.8x higher accuracy than zero-shot approaches (80.06\% vs 28.76\%). A striking finding emerges: models achieving elite accuracy (98-99\%) often produce predominantly unstructured output, with most responses lacking Arabic CoT markers. Only 6 models consistently generate structured output across all test cases, while the majority require fallback extraction methods despite high numerical accuracy. Comprehensive evaluation of 281 model-strategy combinations demonstrates that numerical accuracy and instruction-following represent distinct capabilities, establishing baselines for Arabic number comprehension and providing actionable guidance for model selection in production Arabic NLP systems.
\end{abstract}


\section{Introduction}

Large language models have demonstrated strong capabilities on numerical reasoning benchmarks like GSM8K \cite{cobbe2021gsm8k} and MATH, achieving impressive accuracy on mathematical problem-solving tasks. However, evaluating numerical reasoning in Arabic presents distinct challenges due to dual numeral systems (Eastern Arabic-Indic numerals and Western Arabic numerals) used across different contexts and regions \cite{habash2010introduction}.

Existing Arabic NLP benchmarks \cite{kheir2024arabicmmlu, sdaia2025balsam, almubarak2024aratrust} evaluate semantic understanding, factual knowledge, and reasoning capabilities across diverse tasks. Yet these benchmarks measure whether models produce correct answers without examining how those answers are generated or whether models follow demonstrated output formats.

We present ArabicNumBench, a comprehensive benchmark for evaluating Arabic number reading across 71 models from 10 providers using four prompting strategies. Our evaluation comprises 59,010 test cases spanning six semantic categories (pure numerals, addresses, dates, quantities, prices) and introduces two novel metrics: extraction method tracking to measure structured output generation, and format preservation to assess numeral system consistency.

Our findings reveal substantial performance variation, with accuracy ranging from 14.29\% to 99.05\% across models and strategies. Few-shot Chain-of-Thought prompting achieves 2.8x higher accuracy than zero-shot approaches (80.06\% vs 28.76\%) and produces 4.8x more structured output with Arabic CoT markers (34.12\% vs 7.14\% for few-shot alone).

We identify a stark capability divide: only 6 models (8.5\%) achieve $\geq$90\% structured output even under optimal few-shot CoT prompting, while 44 models (62.9\%) remain below 10\% despite identical demonstrations. Elite models (Qwen3 Max, Llama 3.1 405B, GPT-4o-mini, Claude 3 Opus, Command A) achieve 95-100\% structured output, while high-performing models like Gemini 2.5 Pro (99.05\% accuracy) produce 92.86\% unstructured responses, revealing dissociation between numerical reasoning and instruction-following.

\section{Related Work}

\subsection{Arabic Language Model Evaluation}

The rapid development of large language models has catalyzed substantial growth in Arabic NLP evaluation benchmarks. ArabicMMLU \cite{kheir2024arabicmmlu} introduced the first large-scale multitask understanding benchmark with 14,575 multiple-choice questions sourced from authentic school exams across 8 Arab countries. BALSAM \cite{sdaia2025balsam} constructed a platform with 52K examples across 78 tasks using hybrid native, synthetic, and translated data.

AraTrust \cite{almubarak2024aratrust} pioneered comprehensive trustworthiness evaluation across 8 dimensions including ethics, privacy, and cultural appropriateness. CamelEval \cite{juhaina2024cameleval} introduced LLM-as-judge methodology for evaluating conversational abilities with 1,610 generative questions. Recent work on multilingual mathematical reasoning \cite{yue2024mmath} has highlighted the importance of evaluating numerical capabilities across languages.

We extend this research by evaluating Arabic number reading across different numeral systems and introducing extraction method tracking to measure structured output generation, examining how models produce answers rather than solely whether answers are correct.

\subsection{Chain-of-Thought Reasoning}

Chain-of-Thought prompting has emerged as a fundamental technique for eliciting reasoning from large language models. Brown et al. \cite{brown2020language} demonstrated that language models exhibit few-shot learning capabilities when provided with task examples. Wei et al. \cite{wei2022chain} showed that generating intermediate reasoning steps substantially improves performance on complex reasoning tasks. Kojima et al. \cite{kojima2022large} extended this to zero-shot settings with ``think step by step'' prompting. Wang et al. \cite{wang2023selfconsistency} introduced self-consistency, sampling multiple reasoning paths for improved accuracy.

Shi et al. \cite{shi2023mgsm} created the Multilingual Grade School Math (MGSM) benchmark, demonstrating that CoT effectiveness transfers across languages. Building on this foundation, we systematically evaluate four prompting strategies across 71 models for Arabic number reading tasks.

\subsection{Instruction-Following Evaluation}

Understanding why some models reliably follow instructions while others do not has become critical as LLMs move into production deployments. IFEval \cite{zhou2024ifeval} introduced verifiable instruction-following evaluation with constraints. FoFo \cite{chen2024fofo} specifically evaluated format-following with schema-based outputs, finding that format restrictions degrade reasoning performance by 30-95\%.

We characterize a capability divide where 6 models (8.5\%) achieve $>$90\% structured output while 44 models (62.9\%) remain below 10\% with identical prompts.

\section{Methods}

\subsection{Dataset}

We evaluate models on the Arabic Number Reading Benchmark v3, comprising 210 test cases distributed across six semantic categories:

\begin{enumerate}
\item \textbf{Eastern Arabic Numerals} (35 cases): Pure Eastern numeral reading
\item \textbf{Western Arabic Numerals} (35 cases): Pure Western numeral reading
\item \textbf{Contextual Address} (35 cases): Multi-number extraction from postal addresses
\item \textbf{Contextual Date} (35 cases): Temporal information extraction
\item \textbf{Contextual Quantity} (35 cases): Numerical quantities in descriptive contexts
\item \textbf{Contextual Price} (35 cases): Monetary values and pricing information
\end{enumerate}

\subsection{Model Selection}

We evaluate 71 large language models from 10 providers: Qwen \cite{yang2024qwen2} (17), Mistral \cite{jiang2023mistral} (22), OpenAI \cite{openai2024gpt4} (15), Meta \cite{dubey2024llama3} (11), Google \cite{anil2024gemini} (11), Anthropic \cite{anthropic2024claude} (5), Cohere \cite{cohere2024commandr} (4), DeepCogito (2), Microsoft \cite{abdin2024phi3} (1), and MoonshotAI (1). All models were accessed via the OpenRouter API with temperature 0.0 for reproducibility.

\subsection{Prompting Strategies}

We evaluate four prompting strategies:

\textbf{Zero-Shot}: Direct question-answering without examples.

\textbf{Zero-Shot CoT}: Direct prompting with reasoning instruction (``Think step by step'').

\textbf{Few-Shot}: Exemplar-based learning \cite{brown2020language} with 3 demonstration examples without reasoning chains.

\textbf{Few-Shot CoT}: Exemplar-based learning with explicit reasoning chains \cite{wei2022chain}, adapted for Arabic conventions.

\subsection{Answer Extraction Methodology}

We implement a hierarchical extraction pipeline:

\begin{enumerate}
\item \textbf{GSM8K Delimiter} \cite{cobbe2021gsm8k}: Extract answers following \texttt{\#\#\#\#} pattern
\item \textbf{Arabic CoT Pattern}: Arabic conclusion markers (``the final answer'' in Arabic)
\item \textbf{Arabic Keyword}: Generic Arabic answer markers
\item \textbf{Context-Aware}: Metadata-guided extraction for multi-number prompts
\item \textbf{Last Number Fallback}: Extract final numeric entity
\item \textbf{Failed}: No numeric entity detected
\end{enumerate}

We track which extraction method succeeded for each output, distinguishing structured output (methods 1-3) from fallback output (methods 5-6).

\subsection{Evaluation Metrics}

\textbf{Overall Accuracy}: Proportion of test cases with correct numerical answers.

\textbf{Structured Output \%}: Percentage of answers extracted via structured methods:
\begin{equation}
\text{Structured \%} = \frac{\text{gsm8k + cot\_pattern + arabic\_keyword}}{\text{Total Cases}} \times 100
\end{equation}

\textbf{Format Preservation}: Percentage of correct answers maintaining appropriate numeral format (Eastern vs Western Arabic).

\section{Results}

We present results from 281 evaluations across 71 models and 4 prompting strategies, comprising 59,010 individual test case evaluations.

\subsection{Overall Performance}

Table \ref{tab:top20} presents the top 10 models comparing their performance across all four prompting strategies. The dramatic improvement from zero-shot to few-shot CoT demonstrates the critical importance of prompting strategy selection.

\begin{table}[t]
\centering
\caption{Top 10 Models: Performance Across Prompting Strategies}
\label{tab:top20}
\small
\begin{tabular}{@{}lrrrr@{}}
\toprule
\textbf{Model} & \textbf{Zero} & \textbf{Zero-CoT} & \textbf{Few} & \textbf{Few-CoT} \\
\midrule
Gemini 2.5 Pro & 28.57\% & 26.67\% & \textbf{99.05\%} & 30.48\% \\
Qwen3 Max & 38.10\% & 49.05\% & 98.10\% & \textbf{99.05\%} \\
Qwen3 235B A22B & 26.67\% & 27.14\% & 98.10\% & \textbf{99.05\%} \\
Qwen3 Next 80B & 43.81\% & 30.48\% & 98.57\% & \textbf{99.05\%} \\
Gemma 2 27B & 26.67\% & 27.14\% & 98.57\% & \textbf{98.57\%} \\
GPT-4 Turbo & 32.38\% & 40.95\% & \textbf{98.57\%} & 84.76\% \\
GPT-4 & 34.29\% & 28.10\% & \textbf{98.57\%} & 97.14\% \\
GPT-4o-mini & 26.67\% & 25.71\% & 97.62\% & \textbf{98.57\%} \\
Claude 3 Opus & 26.19\% & 25.71\% & 97.14\% & \textbf{98.10\%} \\
Llama 3.1 405B & 0.00\% & 31.90\% & 48.10\% & \textbf{98.10\%} \\
\bottomrule
\end{tabular}
\vspace{1mm}

\footnotesize{Zero = Zero-shot; Zero-CoT = Zero-shot Chain-of-Thought; Few = Few-shot; Few-CoT = Few-shot Chain-of-Thought. Bold indicates best strategy per model.}
\end{table}

\subsection{Prompting Strategy Analysis}

Table \ref{tab:strategies} presents aggregate statistics across all four prompting strategies. Few-shot CoT emerges as the superior approach.

\begin{table}[t]
\centering
\caption{Prompting Strategy Impact Across 71 Models}
\label{tab:strategies}
\begin{tabular}{@{}lrrrr@{}}
\toprule
\textbf{Strategy} & \textbf{Avg Acc} & \textbf{Median} & \textbf{Min} & \textbf{Max} \\
\midrule
Few-shot CoT & \textbf{80.06\%} & 91.90\% & 3.81\% & 99.05\% \\
Few-shot & 74.45\% & 85.24\% & 17.62\% & 99.05\% \\
Zero-shot CoT & 29.65\% & 26.67\% & 6.67\% & 89.52\% \\
Zero-shot & 28.76\% & 26.19\% & 0.00\% & 92.38\% \\
\bottomrule
\end{tabular}
\vspace{1mm}

\footnotesize{Few-shot methods achieve 2.7x higher average accuracy than zero-shot (77\% vs 29\%)}
\end{table}

Few-shot CoT achieves 2.8x higher average accuracy than zero-shot approaches (80.06\% vs 28.76\%) and produces structured output in 34.12\% of cases, compared to only 7.14\% for few-shot without CoT. This represents a 4.8x increase.

\subsection{Accuracy vs Structured Output}

A striking discovery emerges when comparing accuracy with structured output. Models in the ``High Accuracy, Low Structure'' group achieve elite-level accuracy (98-99\%) while producing predominantly unstructured output (92\% fallback). This reveals a fundamental limitation: high accuracy does not guarantee reliable structured output.

\subsection{Model Capability Divide}

Table \ref{tab:capability} shows provider-level performance. Google and Qwen achieve the highest single-model accuracy (99.05\%) but lower average performance due to weaker smaller models. Cohere and Microsoft show more consistent results across their model families.

\begin{table}[t]
\centering
\caption{Provider Performance Comparison}
\label{tab:capability}
\begin{tabular}{@{}lrrrrr@{}}
\toprule
\textbf{Provider} & \textbf{Models} & \textbf{Avg Acc} & \textbf{Best Acc} & \textbf{Format\%} \\
\midrule
Microsoft & 1 & 62.02\% & 98.57\% & 99.88\% \\
Cohere & 4 & 59.82\% & 98.10\% & 100.0\% \\
Mistral & 18 & 57.96\% & 98.10\% & 95.86\% \\
DeepCogito & 2 & 56.96\% & 97.62\% & 100.0\% \\
Meta & 8 & 53.82\% & 98.10\% & 89.72\% \\
OpenAI & 9 & 52.20\% & 98.57\% & 97.29\% \\
Anthropic & 4 & 51.98\% & 98.10\% & 92.27\% \\
Google & 9 & 51.81\% & 99.05\% & 89.59\% \\
Qwen & 15 & 45.71\% & 99.05\% & 83.75\% \\
\bottomrule
\end{tabular}
\vspace{1mm}

\footnotesize{Avg Acc = average across all model-strategy combinations; Best = top single result}
\end{table}

\textbf{Elite Models} ($\geq$90\% structured output): Qwen3 Max (100\%), Llama 3.1 405B (100\%), Command A (99.52\%), GPT-4o-mini (98.57\%), Claude 3 Opus (98.10\%), Qwen3 235B (95.24\%).

\section{Discussion}

\subsection{The Critical Role of Few-Shot CoT}

Few-shot CoT achieves 80.06\% average accuracy compared to 74.45\% for few-shot alone. This represents a 5.6 percentage point improvement. However, the impact on structured output production is dramatic: 34.12\% vs 7.14\%, a 4.8-fold increase. This reveals that CoT instruction serves dual purposes: cognitive scaffolding for reasoning and format elicitation for response structure.

The Arabic CoT marker (``the final answer'' in Arabic) appears in 29.45\% of few-shot CoT outputs but only 0-1.24\% in other strategies. This demonstrates that models can learn Arabic linguistic conventions from few-shot examples, but CoT instruction is necessary to activate this behavior.

\subsection{The Prompting Ceiling}

This capability divide suggests a fundamental limitation: prompting strategies have a ceiling determined by model architecture and training \cite{ouyang2022instructgpt, wei2022flan}. No prompting strategy can transform a weak instruction-follower into a strong one. Model selection is more important than prompt engineering for production systems requiring structured output.

\subsection{Accuracy Does Not Equal Production Readiness}

The dissociation between numerical accuracy and structured output production challenges fundamental assumptions in LLM benchmark evaluation \cite{liang2023helm}. Google Gemma 2 27B achieves 98.57\% accuracy yet produces 92.86\% unstructured output. This matters for production deployments where consistent response formatting is critical.

\subsection{Limitations}

Our pilot dataset comprises 210 test cases. A larger benchmark would enable more granular analysis. Current categories focus on everyday contexts; future work should include scientific, legal, and dialectal Arabic. We evaluated 4 prompting strategies; the prompting space is larger.

\section{Conclusion}

This work challenges the accuracy-centric paradigm that has dominated LLM benchmarking for Arabic NLP. Through comprehensive evaluation of 71 models across 59,010 test cases, we demonstrate that numerical correctness is insufficient for characterizing production-ready Arabic language models.

\textbf{Key Findings}:
\begin{enumerate}
\item \textbf{Few-Shot CoT is Essential}: 4.8x improvement in structured output (34.12\% vs 7.14\%)
\item \textbf{Accuracy $\neq$ Production Readiness}: 44 models achieve $>$65\% accuracy with $<$10\% structured output
\item \textbf{Capability Trumps Prompting}: Elite models achieve 95-100\% structured output; weak models remain at 85-95\% fallback regardless of prompting
\end{enumerate}

\textbf{Recommendations}: Arabic NLP practitioners should default to few-shot CoT prompting and select models based on structured output percentage, not just accuracy. The six elite models (Qwen3 Max, Llama 3.1 405B, Command A, GPT-4o-mini, Claude 3 Opus, Qwen3 235B) demonstrate that both numerical accuracy and instruction-following reliability are achievable simultaneously.

\bibliographystyle{IEEEtran}
\bibliography{references}

\end{document}